\documentclass[10pt,twocolumn,letterpaper]{article}

\usepackage{wacv}
\usepackage{times}
\usepackage{epsfig}
\usepackage{graphicx}
\usepackage{amsmath}
\usepackage{amssymb}
\usepackage{textcmds}
\usepackage{authblk}

\usepackage[pagebackref=true,breaklinks=true,letterpaper=true,colorlinks,bookmarks=false]{hyperref}

\wacvfinalcopy 

\def\wacvPaperID{1243} 

\usepackage[ruled,vlined]{algorithm2e}
\ifwacvfinal\fi
\begin{document}

\title{Multi-Modal Association based Grouping for Form Structure Extraction}



\author[1]{Milan Aggarwal}
\author[1]{Mausoom Sarkar}
\author[2]{Hiresh Gupta}
\author[1]{Balaji Krishnamurthy}
\affil[1]{Media and Data Science Research Lab, Adobe}
\affil[2]{Adobe Experience Cloud}


\maketitle
\ifwacvfinal\thispagestyle{empty}\fi

\begin{abstract}

Document structure extraction has been a widely researched area for decades. Recent work in this direction has been deep learning-based, mostly focusing on extracting structure using fully convolution NN through semantic segmentation. In this work, we present a novel multi-modal approach for form structure extraction. Given simple elements such as textruns and widgets, we extract higher-order structures such as TextBlocks, Text Fields, Choice Fields, and Choice Groups, which are essential for information collection in forms. To achieve this, we obtain a local image patch around each low-level element (reference) by identifying candidate elements closest to it. We process textual and spatial representation of candidates sequentially through a BiLSTM to obtain context-aware representations and fuse them with image patch features obtained by processing it through a CNN. Subsequently, the sequential decoder takes this fused feature vector to predict the association type between reference and candidates. These predicted associations are utilized to determine larger structures through connected components analysis. Experimental results show the effectiveness of our approach achieving a recall of $90.29\%$, $73.80\%$, $83.12\%$, and $52.72\%$ for the above structures, respectively, outperforming semantic segmentation baselines significantly. We show the efficacy of our method through ablations, comparing it against using individual modalities. We also introduce our new rich human-annotated Forms Dataset.
\end{abstract}

\section{Introduction}

\begin{figure*}[t]
\centering
\includegraphics[width=14cm,height=5cm]{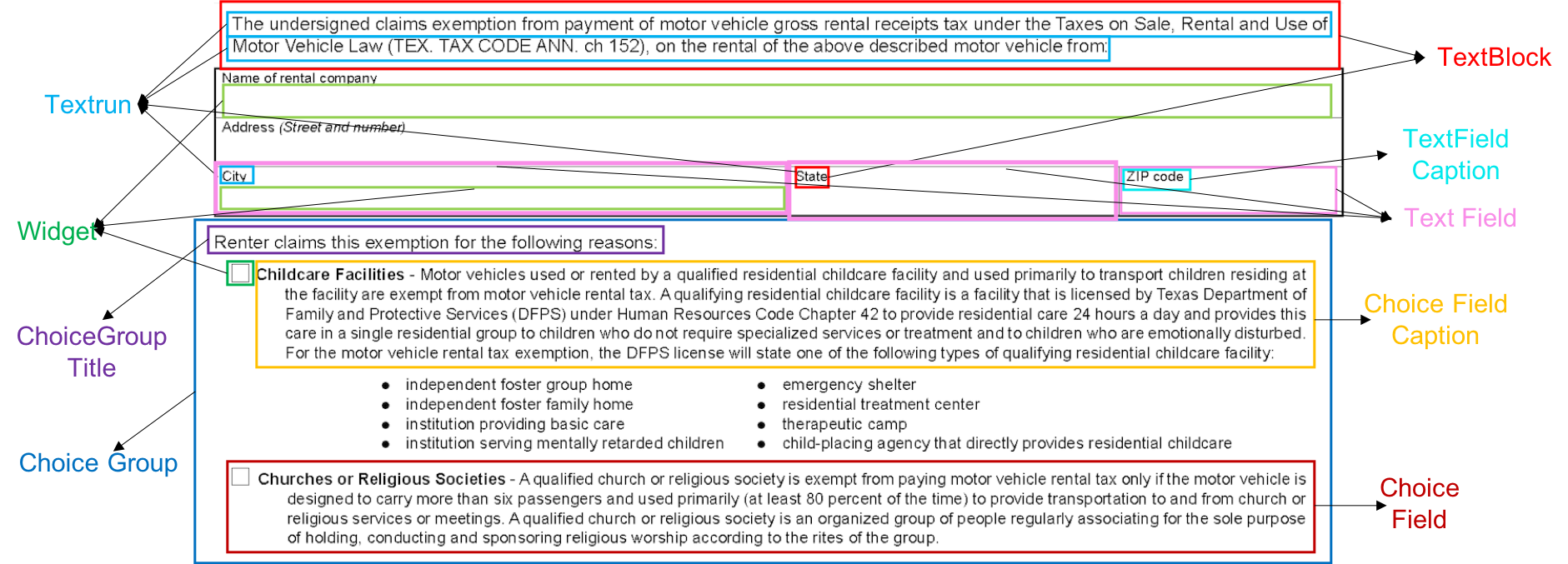}
\vspace{-4pt}
\caption{Part of an example form illustrating elements and structures at different levels of hierarchy.}
\label{fig:describe}
\end{figure*}


Document semantic structure extraction (DSSE) is an actively-researched area \cite{tablerule,tableprice,historical,colorado}. Forms are documents that are widely used to acquire information from humans across different domains such as health care, administration and financial systems. Organizations using forms for several decades would like to digitize them to enable indexing, accessibility etc. for themselves as well as for providing seamless digital experiences (such as re-flow) to their users \cite{devices,mobile,dictionary,caseHTML}. In order to do this, forms are converted into PDFs/Acroforms with information about only elementary level structures like text,images,widgets embedded in them. However, they do not contain any information about higher order structures. Since the elementary level structural information is not enough to digitise a form document, there is a need to extract higher order constructs.


A form page broadly comprises of textruns and widgets as its smallest constituent elements. Widgets are spaces provided to fill information. Every widget has some text associated with it which describes what needs to be filled in the widget. A textrun is a small chunk of text, often a single line or a word. Such lower elements act as building blocks in larger constructs such as textblocks, text fields, choice fields, choice groups etc. A textblock is a logical block of self-contained text comprising of one or more textruns, a text field is a collection of a caption and widgets where caption is a textblock which describes what to fill in the widgets. A choice field is a boolean field containing a caption, a radio button type widget and optional widgets provided to write text if required. A ChoiceGroup is a collection of choice fields and an optional choice group title which is a textblock that specifies various details or instructions regarding filling the choice fields. Figure \ref{fig:describe} shows different elements and structures in a sample form.

In this work, we propose a hierarchical multi-modal bottom up approach to detect larger constructs in a form page. We create a two step pipeline where we train a model to predict textblocks given textruns and widgets as input in the first step and train another model (having similar architecture) separately in the second step to extract larger constructs that are intrinsic and unique to forms such as text fields, choice fields and choice groups using the widgets and extracted textblocks. We leverage the fact that elements constituting a larger construct are present in close proximity. For instance, textruns constituting a textblock can be consecutive lines of text. Similarly, choice fields that are part of a choice group are arranged closer to one another. 

Specifically, we determine a local neighbourhood around each element (reference) based on fixed number of other elements (candidates) nearest to it (candidates include the reference as well for guidance). We train our models to identify candidates in the neighbourhood with which reference can be associated as part of a larger construct. To achieve this, we extract an image patch corresponding to the reference neighborhood and process it through a CNN. We also extract text content of the candidates and process it through an LSTM \cite{hochreiter1997long} based text encoder to obtain their textual representation. Additionally, we concatenate spatial coordinates of candidates with their textual representation. For a given candidate, since the association with the reference can be guided depending on other candidates, we arrange the candidates in a sequence according to natural reading order. We process the corresponding sequence of concatenated representations through a BiLSTM \cite{huang2015bidirectional} to obtain context aware candidate feature representations. Subsequently, we use an attention based fusion mechanism \cite{sinha2019attention} to combine visual, textual and spatial modalities by fusing context aware feature representation of each candidate with the output of CNN. The fused representation corresponding is passed through a final Seq2Seq \cite{sutskever2014sequence} based decoder that sequentially predicts association between the reference and different candidate elements. These associations are used to construct an association graph between different elements such that final higher order elements are extracted through determining connected components. 

Document structure extraction has been studied extensively with recent works mostly employing deep learning based fully convolution neural networks \cite{tableprice,historical,colorado}. They extract structures through performing semantic segmentation \cite{FCNSS} over document image. Such methods perform well at extracting coarser structures. However, our Forms Dataset comprises of dense form images (having resolution up to 2000 pixels) with closely spaced elements and higher order constructs. Segmentation based methods generally extract structures for the entire page in a single forward pass. Due to memory constraints, such methods downscale the original image before giving it as input to their model. Hence they end up merging different structures since down-scaling makes it difficult to disambiguate closely spaced structures. We instead take a different approach where we formulate structure extraction as an association task such that these associations are made based on local contextual information. Our contributions can be listed as :

\begin{itemize}
    \item We propose a multi-modal approach for forms structure extraction, specifically for the task of extracting higher order constructs from lower level elements. We perform ablations by evaluating individual modalities for the same task and comparing against the multi-modal version.
    \item We show the effectiveness of our approach to disambiguate closely spaced structures. Our approach outperforms two semantic segmentation baselines designed for segmenting natural and document images respectively.
    \item We introduce our new rich human annotated Forms Images Dataset comprising of text content, spatial information tags and other larger construct annotations.
\end{itemize}

\section{Related Work}
A closely related  problem is of document structure extraction. Earlier works for document layout analysis have mostly been rule based with hand crafted features for extracting coarser structures such as graphics and text paragraphs \cite{rule1}, cut algorithms \& connected components for segmenting image and determining text regions \cite{rule3}, and other methods for determining physical layouts \cite{rule4}. These are generally categorized into bottom-up \cite{rule5} and top-down approaches \cite{rule2} with former focusing on detecting text lines \& composing them into paragraphs and the latter detect page layout by diving it into columns and blocks.

With the advancement in deep learning, recent approaches have mostly been fully convolution neural network (FCN) based that eliminate the need of designing complex heuristics \cite{tableprice,historical,colorado}. FCNs were successfully trained for semantic segmentation \cite{FCNSS} which has now become a common technique for page segmentation. The high level feature representations make FCN effective for pixel-wise prediction. FCN has been used to locate/recognize handwritten annotations, particularly in historical documents \cite{handwriting}. \cite{curtis} proposed a model that jointly learns handwritten text detection and recognition using a region proposal network that detects text start positions and a line following module that incrementally predicts the text line which is subsequently used for reading.

Several methods have addressed regions in documents other than text such as tables, figures etc. Initial deep learning work that achieved success in table detection relied on selecting table like regions on basis of loose rules which are subsequently filtered by a CNN \cite{tablerule}. \cite{tableprice} proposed multi-scale, multi-task FCN comprising of two branches to detect contours in addition to page segmentation output that included tables. They additionally use CRF (Conditional Random Field) to make the segmented output smoother and induce dependency among pixels. For the task of figure extraction from scientific documents, \cite{figures} introduced a large scale dataset comprising of 5.5 million document labels. They find bounding boxes for figures in PDF by training Overfeat \cite{overfeat} on image embeddings generated using ResNet-101 on this dataset.

Few works have explored alternate input modalities such as text for other document related tasks. Multi-modal methods have been proposed to extract predefined named entities from invoices \cite{katti2018chargrid,liu2019graph}. Graph Neural Networks (GNNs) have been used to detect tables in invoice documents \cite{ribatable} as well as for association based table structure parsing \cite{qasim2019rethinking}. Document classification is a partly related problem that has been studied using CNN-only approaches for identity document verification \cite{sicre2017identity}. \cite{HAN} have designed HAN which hierarchically builds sentence embeddings and then document representation using multi-level attention mechanism. Other works explored multi-modal approaches, using MobileNet \cite{mobilenets} and FastText \cite{fasttext} to extract visual and text features respectively, which are combined in different ways (such as concatenation) for document classification \cite{docclassmulti}. In contrast, we tackle a different task of form layout extraction which require recognising different structures present in a form document.

\cite{colorado} also proposed a multimodal FCN (MFCN) to segment figures, tables, lists etc. in addition to paragraphs from documents. They concatenate a text embedding map to feature volume according to pixel-text correspondence in the original image. In addition, they show improvements through unsupervised learning using region consistency loss over the segmented output to obtain regular segmentation. As baselines for the tasks proposed, we consider image based semantic segmentation approaches. We compare the performance of our approach with 1) their FCN based method and 2) DeepLabV3+ \cite{deeplabv3plus2018}, which is state of the art deep learning model for semantic segmentation.

\begin{figure*}[t]
\centering
\includegraphics[width=12.5cm,height=3.5cm]{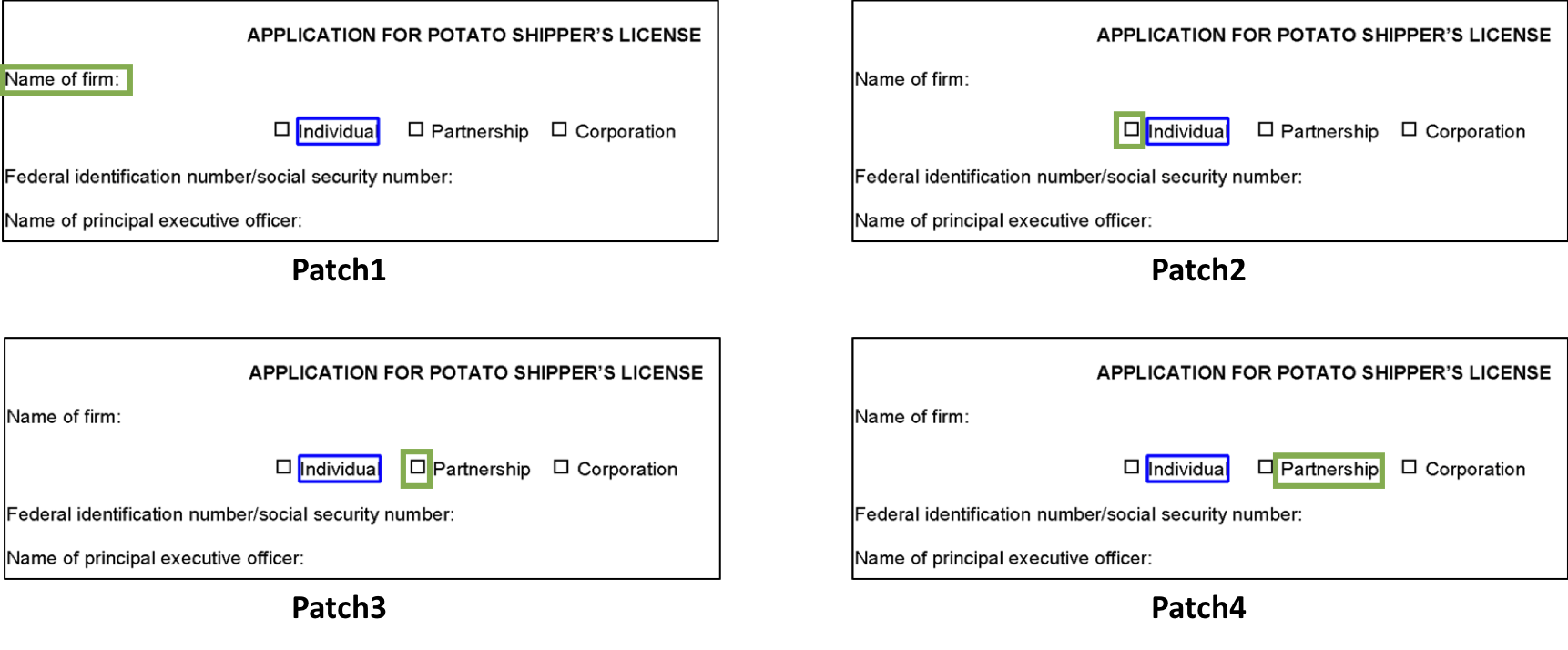}
\vspace{-4pt}
\caption{Sub-sequence of patch images for reference textblock "Individual" (blue): Patch images corresponding to candidates (green) in sub-sequence have been shown.}
\label{fig_patches}
\end{figure*}


\section{Methodology}
In this section, we discuss our approach to detect various elements like textblocks, text fields, choice fields and choice groups in a form document. Since each of these higher order constructs is a collection of lower elements, we formulate their detection as a grouping task over such elements. We achieve this through a two step pipeline where in the first step, we train a model that groups textruns into textblocks given textruns and widgets as input. And in the second step, we train another model separately having similar architecture but which takes textblocks and widgets as inputs and groups them into higher order constructs - text fields, choice fields and choice groups.

\subsection{Input Pipeline}
Let $tr$, $tb$ and $wi$ be the set of textruns, textblocks and widgets respectively present in a form page. For both the pipeline steps, let $t_1$ and $t_2$ be the collection of lower elements. Here, $t_1 = tr$ and $t_2 = wi$ for textblock prediction step while $t_1=tb$ and $t_2=wi$ for higher order construct grouping. For every element $e_r\in t_1$, we determine a neighbourhood patch $p$ around it comprising of elements $e^p$ which are obtained by ranking elements in $t_1 \bigcup t_2$ in ascending order according to their distance from $e_r$. We then pick top $k_1$ and $k_2$ elements from $t_1$ and $t_2$ respectively according to the ranked list. We call $e_r$ as reference element while other selected elements as candidate elements. Hence, $e^p = e_{1}^p \bigcup e_{2}^p$; $e_{1}^p \subseteq t_1$; $e_{2}^p \subseteq t_2$; $\vert$$e_{1}^p$$\vert$ $\leq k_1$; and $\vert$$e_{2}^p$$\vert$ $\leq k_2$. Each form page element's location is specified through four bounding box values - $x$, $y$, $w$, $h$, where $x$ and $y$ are the coordinates of mid-point of bounding box  and $w$ and $h$ are its width and height respectively. Given reference element $a$ and candidate element $b$, distance of $b$ form $a$ is
\begin{equation}
\label{eqn_loss}
\begin{split}
d(a,b) = 10 * min(|y_a - (y_b-h_{b}/2)|, \\ |y_a - y_b|, |y_a - (y_b+h_{b}/2)|)  + \\ 
min(|x_a - (x_b-w_{b}/2)|, \\ |x_a - x_b|, |x_a - (x_b+w_{b}/2)|)  
\end{split}
\end{equation}




We consider elements which are present closer in vertical direction to the reference to be more appropriate choice for candidate elements. This takes more horizontal context into account (hence a larger weight of 10 to vertical distance term in eq\eqref{eqn_loss}). We measure the distance between reference's mid point with the middle and extreme edges of candidate elements from both the directions leading to our distance formulation above. Once $e^p$ is determined, we cut an image patch $p$ around the reference $e_r$ which is determined through taking union of bounding boxes of elements in $e^p$. We highlight the reference $e_r$ in $p$ by drawing its bounding box in blue. We arrange the elements in $e^p$ according to natural reading order (from left to right and top to bottom) to obtain a sequence of elements $a^p = \{a^p_1, a^p_2, a^p_3, ..., a^p_{k_1+k_2}\}$. Let ${bb}^a = \{{bb}^a_1, {bb}^a_2, ..., {bb}^a_{k_1+k_2}\}$ be the corresponding bounding box coordinates sequence such that ${bb}^a_i = \{x^a_i,y^a_i,w^a_i,h^a_i\}$ represent x-y coordinates of the top left corner, width and height respectively. We normalise these coordinates to range $[0,1]$ with respect to the unified bounding box of the patch (through considering top left corner of patch as origin and dividing by patch dimensions), thus obtaining ${bb}^n = \{{bb}^n_1, {bb}^n_2, ..., {bb}^n_{k_1+k_2}\}$. For each element $a^p_i$, we modify $p$ to highlight $a^p_i$ by drawing its bounding box around it in green to create an image patch ${im}^p_i$ like in fig\ref{fig_patches}. We then re-size the modified patch image to $H \times W$ and concatenate a normalised 2-D mesh grid of same resolution along the channels, obtaining a 5-channel image. We thus obtain the corresponding final patch image sequence ${im}^f_p = \{{im}^p_1, {im}^p_2, {im}^p_3, ..., {im}^p_{k_1+k_2}\}$ where each patch has one element $a^p_i$ highlighted. In addition to these images we also have text content $t^a_i$ of each element $a^p_i$ . The sequences ${bb}^n$, ${im}^f_p$ and $t^a$ are given as input to the model that predicts which elements can be associated with the reference element $e_r$ to form a higher order group.

\subsection{Model Architecture}
\begin{figure*}
\centering
\includegraphics[width=15.5cm,height=8.5cm]{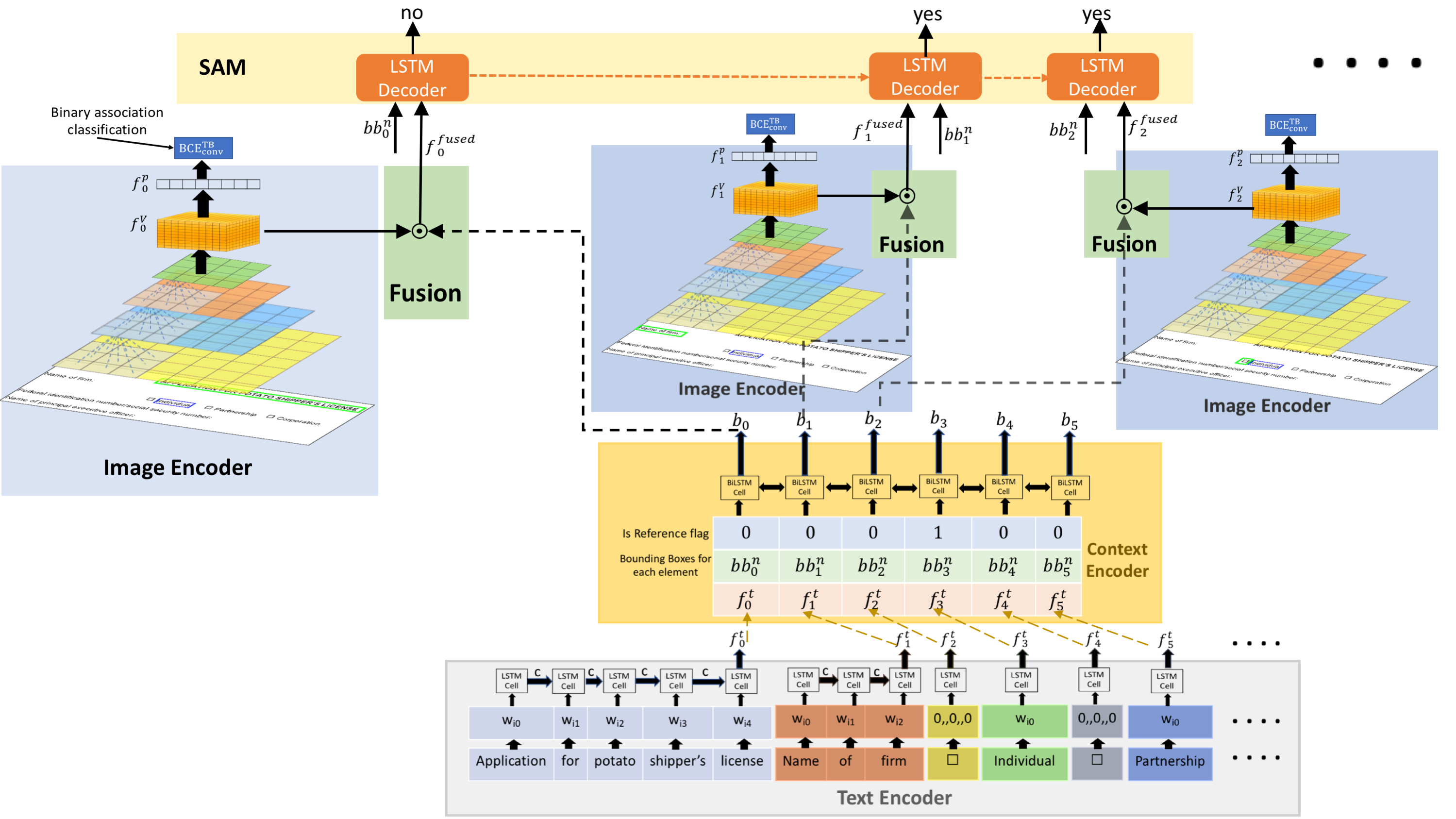}
\vspace{-4pt}
\caption{Detailed view of our model architecture with input patches for different candidates given as input to CNN. Spatial and textual information is processed through Text Encoder (TE) and Context Encoder (CE). This representation is fused with CNN output for each candidate. Fused representation is given as input to Association Decoder that predicts the final output.}
\label{complete_model}
\end{figure*}

Our model comprises of different sub-modules trained jointly - Image Encoder (IE) that takes each patch image in ${im}^f_p$ to generate its feature representation through a CNN; LSTM based Text Encoder (TE) that processes each $t^a_i$ to obtain its text embedding; a Bi-LSTM based Context Encoder (CE) that takes the concatenated sequence of normalised spatial coordinate and text embedding to generate context aware representations for each candidate in the patch, Fusion Module (FM) that fuses these context aware representations with the corresponding image patch features, and a final LSTM based Sequential Association Module (SAM) that predicts the association between reference element $e_r$ and other candidate elements in the sequence. Figure \ref{complete_model} shows a detailed architecture of our approach depicting each sub-module and overall model pipeline. We discuss each of these modules in detail.

\textbf{Image Encoder : } Each image patch ${im}^p_i$ in the sequence ${im}^f_p$ has size $H \times W \times 5$ (RGB patch with 2d mesh grid) and is processed through a CNN comprising of $n_{cb}$ convolution blocks. Convolution block ${cb}_j$ comprises of $n^{cl}_j$ convolution layers each having $f_j$ filters with kernel size $k_j \times k_j$, stride $=$ 1. We discuss the exact value of the hyper-parameters later in implementation details. After each convolution block, we apply a max-pooling layer having kernel size $3 \times 3$ and stride $=$ 2. CNN outputs a feature representation $f^V_i$ of size $H' \times W' \times C'$ where $C'$ is the number of feature maps. This $f^V_i$ is later on used in the multi modal fusion phase. In addition to this an auxiliary branch is created with a flattened $f^V_i$ being passed through a FCN comprising of 2 fully connected layers each having ${FC_c}$ units to obtain the final representation $f^p_i$ for image patch ${im}^p_i$. Finally, $f^p_i$ is passed through the final prediction FC layers which predicts the association of the current candidate element $a^p_i$ with the reference element. For the first pipeline step where we group textruns into textblocks, we use a single FC prediction layer with sigmoid activation which gives a binary classification output; while for second pipeline step, we use two FC prediction layers - one with softmax activation to perform 3-way classification to predict if a textblock can be grouped with a candidate textblock or widget into \qq{field}, \qq{choicefield} or \qq{not related} while the second prediction layer outputs a binary decision if a textblock can be grouped with a candidate textblock or widget into a choice group. Using the auxiliary branch helps in training the CNN features which are used further in fusion stage and Sequential Association Module.

\textbf{Text Encoder :} Consider an element $a^p_i$ having text $t^a_i$ comprising of words $\{w_{i1},w_{i2}, ..., w_{in}\}$. Since the text information is obtained through PDF content, the words often contain noise, making use of standard word vectors difficult. To mitigate this, we obtain word embeddings using python library \textit{chars2vec}\footnote{https://github.com/IntuitionEngineeringTeam/chars2vec}. This gives a sequence of embeddings $\{we_{i1}, we_{i2}, ..., we_{in}\}$ which is given as input to an LSTM - $TE$, that processes the word embeddings such that the cell state $c^t_i$ after processing last word is used as text representation for $a^p_i$. A widget's textual representation is taken as a vector of 0s. This text representation is further passed through a fully connected layer with 100 units and ReLU activation to obtain $f^t_i$. 

\textbf{Context Encoder :} Consider a sequence element $a^p_i$ with corresponding textual representation $f^t_i$ and spatial coordinates ${bb}^n_i$. Let $r_i$ be a binary flag indicating if the $i^{th}$ element in the sequence is the reference element. Hence it is 1 for reference element and 0 for others. We obtain $ce_{i} = [{bb}^n_i||f^t_i||r_i]$ to represent the element, where operator $||$ represents concatenation. The sequence $ce$ is given as input to a BiLSTM \cite{huang2015bidirectional} - $CE$, which produces a context aware representation $b_i$ for each element in the sequence. 

\textbf{Fusion Module :}  In this module, we fuse the context aware representations $b_i$ with the corresponding image patch feature representation $f^V_i$ following attention based fusion mechanism \cite{sinha2019attention}. To achieve this, we use $b_i$ as $1 \times 1$ filter to perform 2-D convolution over feature volume $f^V_i$. To achieve this we configure the size of LSTMs in CE to make it compatible with $C'$. This gives a single channel fused feature map with dimension $H' \times W'$. We flatten this fused map to obtain $f^{fused}_i$ having $H*W$ dimension which is used in the final decoder stage. 

\textbf{Sequential Association Module :} Our SAM module comprises of standard Seq2Seq \cite{sutskever2014sequence} decoder where we predict association decision between reference and candidate $a^p_i$ sequentially conditioned on predictions made for previous candidates $a^p_{j}$ where $j<i$. To achieve this, we use LSTM based decoder $SAM_{LSTM}$ which is given as input $[{bb}^n_i||f^{fused}_i||pred_{i-1}]$, where $pred_{i-1}$ is prediction made for the previous candidate (as in Seq2Seq framework). Additionally, we use Bahdnau attention mechanism \cite{bahdanau2014neural} to make $SAM_{LSTM}$ attend on context memory M, where M is obtained by stacking CE outputs $\{b_1;b_2; ..., b_{k_1+k_2}\}$ column-wise. We use fully connected prediction layers over the outputs of $SAM_{LSTM}$ for both pipeline steps similar to the prediction layers used in the auxiliary branch of IE.
\subsection{Training and Inference Details}
All the modules - IE, TE, CE, FM and SAM are trained end-to-end. In the first pipeline $trw2tb$, we train the network to predict association among the textruns that constitute a textblock. Patch information determined corresponding to a reference textrun constitutes a single training sample for our model. We use binary cross entropy loss (BCE) over binary association predictions made by auxiliary branch of IE ($BCE^{TB}_{conv}$) and sequential predictions made by SAM ($BCE^{TB}_{seq}$) to train entire network. Hence, 

\begin{equation}
    loss_{trw2tb} = BCE^{TB}_{conv} + BCE^{TB}_{seq}
\end{equation}

Similarly, the second pipeline $tbw2fcg$ (having similar architecture but trained separately) is trained to predict associations among textblocks and widgets that constitute fields/choice fields and choice groups through separate prediction layers discussed above. For field and choice field association task, we train the network to associate textblock to other textblocks and widgets within its local patch that are part of the same construct and predict construct type i.e. field or choice field through 3-way classification. Given a choice group $chgp$, consider a reference $e_r$ and a candidate $e_c$ in its patch. Let $e^{CGT}_r = 1$ if $e_r$ is the title of $chgp$, $e^{CFC}_r = 1$ if $e_r$ is the caption of a choice field in $chgp$, $e^{CGT}_c = 1$ if $e_c$ is the title of $chgp$, $e^{CFC}_c = 1$ if $e_c$ is the caption of a choice field in $chgp$ and $e^{CW}_c = 1$ if $e_c$ is a widget of a choice field in $chgp$ and $e_r$ is the caption of same choice field. Let $e^{CGT}_r$, $e^{CFC}_r$, $e^{CGT}_c$, $e^{CFC}_c$, $e^{CW}_c$ be $0$ otherwise. We create training label $label^r_c$ for choice group task as follows :

\begin{center}
$
label^r_c =
\left\{
	\begin{array}{ll}
		1 & \mbox{if } e^{CGT}_r = 1 \ and \  e^{CFC}_c = 1 \\
		1 & \mbox{if } e^{CFC}_r = 1 \ and \ e^{CGT}_c = 1 \\
		1 & \mbox{if } e^{CFC}_r = 1 \ and \ e^{CW}_c = 1  \\
		1 & \mbox{if } e^{CFC}_r = 1 \ and \ e^{CFC}_c = 1  \\
		0 &  otherwise
	\end{array}
\right.
$    
\end{center}

We create the positive labels selectively since associating a choice group title with the widgets of its choice fields might end up confusing the network. We compute cross entropy loss (CE) over field classification predicted by auxiliary branch of IE ($CE^{Field}_{conv}$) and sequential predictions made by SAM ($CE^{Field}_{seq}$) and binary cross entropy (BCE) loss over choice group associations predicted by IE ($BCE^{Chgp}_{conv}$) and SAM ($BCE^{Chgp}_{seq}$) to train entire network. Hence, total loss for pipeline $tbw2fcg$ model becomes

\begin{equation}
\begin{split}
    loss_{tbw2fcg} = CE^{Field}_{conv} + CE^{Field}_{seq} + \\ BCE^{Chgp}_{conv} + BCE^{Chgp}_{seq}
\end{split}    
\end{equation}

We train the $tbw2fcg$ model on tagged textblocks and widgets data. During inference, we use the textblocks predicted by the $tr2tb$ model with ground truth widgets as input for evaluation. Additionally, we use ground truth labels as previous step prediction input for $SAM$ during training and use its own prediction during inference following standard teacher forcing technique \cite{williams1989learning}. \\

\noindent \textbf{Post Processing and Evaluation Metric} : 
Since our model predicts associations between different elements, we obtain the final construct by finding connected components over element association graph. For textblock prediction, we construct an undirected graph $G_{TB} = (tr,E)$ such that an edge between ${tr}_i$ and ${tr}_j$ exists iff the network associates reference textrun ${tr}_i$ with candidate textrun ${tr}_j$ and vice-versa. Likewise, we follow a similar approach for determining fields, choice fields and choice groups. Since textruns and widgets are building blocks for every higher order construct, we decompose a higher order construct into its constituent textruns and widgets to determine correct matches. Given a set of tagged groups $\{g_1,g_2,g_3, ... ,g_m\}$ and a set of predicted groups $\{p_1,p_2,p_3, ... ,p_k\}$ of the same construct type, we say a group $p_i$ matches $g_j$ iff the former contains exactly the same Textruns and Widgets as the latter. This is a stricter measure compared to image based methods which mostly uses IoU.

\section{Experiments}



\subsection{Dataset}

We have used our Forms Dataset\footnote{We plan to make a part of this dataset available at https://github.com/MMPAN-forms/MMPAN} comprising of $~5K$ forms. Textruns' text content and bounding box coordinates are obtained from the PDF used for extracting form page images. We got the dataset annotated with bigger constructs such as text fields, choice fields and choice groups. We take a split of 4417 and 490 form pages for training and testing respectively. Table \ref{table:data} summarises dataset statistics.

\begin{table}[h]
 \centering
 \caption{Dataset Statistics}
 \label{table:data}
 \begin{tabular}{|c|c|c|}

 \hline 
  Construct & \# Train & \# Test \\
  \hline 
  TextBlocks & 192188 & 21141 \\
  \hline
  Text Fields & 65789 & 6678 \\
  \hline
  Choice Fields & 43959 & 4426 \\
  \hline
  ChoiceGroups & 15507 & 1762 \\
  \hline
\end{tabular}
\end{table}

\begin{table*}[t]
 \centering
 \caption{Comparison of Recall and Precision (in \%) for different structures of our model with the baselines and ablation methods}
 \label{Results:pipeline}
 \begin{tabular}{|c|c|c|c|c|c|c|c|c|}
 \hline 
  \textbf{Method} & \multicolumn{2}{|c|}{\textbf{TextBlock}} & \multicolumn{2}{|c|}{\textbf{TextField}} & \multicolumn{2}{|c|}{\textbf{ChoiceField}} & \multicolumn{2}{|c|}{\textbf{ChoiceGroup}}  \\
  \hline
   & Recall & Precision & Recall & Precision & Recall & Precision & Recall & Precision \\
  
  \hline
  $DLV3+$ & 22.10 & 42.39 & 19.35 & 30.87 & 21.46 & 25.49 & 38.06 & 23.16\\
  $MFCN$ & 63.25 & 37.23 & 37.12 & 38.94 & 31.45 & 14.24 & 30.99 & 26.85\\
  $wo TE$ & 88.62 & 85.74 & 68.97 & 78.81 & 81.65 & 85.82 & 46.93 & 44.29 \\ 
  $wo CE$ & 89.06 & 85.64 & 70.76 & \textbf{81.28} & 82.58 & \textbf{86.14} & 46.99 & 45.67 \\
  $CPAN$ & 87.96 & 85.83 & 69.88 & 78.56 & 81.54 & 83.08 & 42.73 & 38.87\\
  $SPAN$ & 84.96 & 74.45 & 60.69 & 66.50 & 72.72 & 79.11 & 34.90 & 31.79\\
  $MMPAN$ & \textbf{90.29} & \textbf{85.94 }& \textbf{73.80} & 80.01 & \textbf{83.12} & 85.91 & \textbf{52.72} & \textbf{52.22}\\
  \hline
\end{tabular}
\end{table*}

\begin{table*}[t]
 \centering
 \caption{Recall and Precision (in \%) of our proposed approach and ablation methods. Tagged textblocks are given as input to second pipeline step instead of first step outputs.}
 \label{Results:gt}
 \begin{tabular}{|c|c|c|c|c|c|c|c|c|}
 \hline 
  \textbf{Method} & \multicolumn{2}{|c|}{\textbf{TextBlock}} & \multicolumn{2}{|c|}{\textbf{TextField}} & \multicolumn{2}{|c|}{\textbf{ChoiceField}} & \multicolumn{2}{|c|}{\textbf{ChoiceGroup}}  \\
  \hline
   & Recall & Precision & Recall & Precision & Recall & Precision & Recall & Precision \\
  \hline
  $woTE_{GT}$ & 88.62 & 85.74 & 82.40 & 89.80 & 88.16 & 90.35 & 52.04 & 48.98 \\
  $woCE_{GT}$ & 89.06 & 85.64 & 84.37 & \textbf{92.43} & 89.63 & 90.68 & 54.99 & 51.62 \\
  $CPAN_{GT}$ & 87.96 & 85.83 & 83.68 & 89.47 & 88.93 & 88.65 & 48.52 & 44.07\\
  $SPAN_{GT}$ & 84.96 & 74.45 & 72.63 & 79.66 & 79.95 & 85.40 & 44.55 & 40.27\\
  $MMPAN_{GT}$ & \textbf{90.29} & \textbf{85.94 }& \textbf{86.83} & 90.87 & \textbf{89.70} & \textbf{90.70} & \textbf{60.15} & \textbf{57.73}\\
  \hline
\end{tabular}
\end{table*}

\subsection{Implementation Details}
In the input pipeline, we set large enough values of $k_1=6$ and $k_2=4$ for textblock prediction and $k_1=10$ and $k_2=4$ for the second pipeline model. For Image Encoder IE, we set input resolution $H = 160$ and $W = 640$, $n_{cb} = 5$ with $[n^{cl}_j]_{j=1,2,3,4,5} = [2,2,3,3,3]$, $[f_j]_{j=1,2,3,4,5}=[32,64,96,128,256]$ and $[k_j]_{j=1,2,3,4,5}=[5,3,3,3,3]$. CNN output feature volume has dimensions $H'=5$, $W'=20$ and $C'=256$. In auxiliary branch, ${FC}_C=1024$. All convolution layers have ReLU activation. For the text encoder $TE$, we fix the size of text in an element to a maximum of 200 words. We use chars2vec model which outputs 100 dimensional embedding for each word and fix the LSTM size to 100. For context encoder CE, we use a hidden size of $128$ for both forward and backward LSTMs. We set the hidden size of $SAM_{LSTM}$ to 1000 with size of attention layer kept at 500. We train all the models using Adam Optimizer \cite{kingma2014adam} at learning rate $1 \times 10^{-4}$ and batch size 8 without hyper-parameter tuning on a single Nvidia 1080Ti GPU.

\begin{figure*}
\centering
\includegraphics[width=13cm,height=12.5cm]{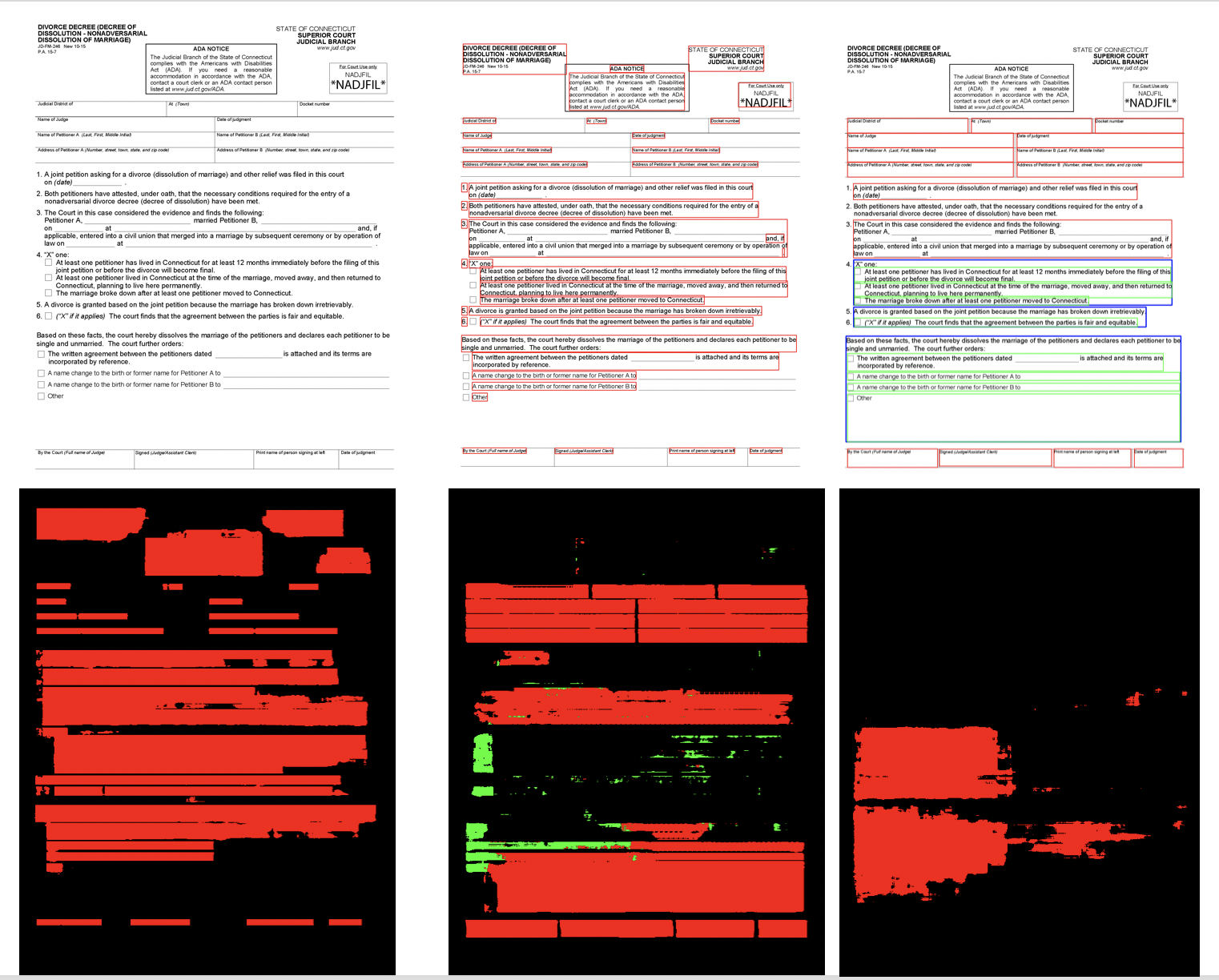}
\vspace{-4pt}
\caption{Visualization : Top row shows sample form image (left), bounding box of textblocks (middle - red), text fields (right - red), choice fields (right - green) and choice groups (right - blue) predicted by our method. Bottom row shows MFCN segmentation masks - textblocks (left), text fields (middle - red), choice fields (middle - green) and choice groups (right).}
\label{visualization}
\end{figure*}

\subsection{Results}

\noindent \textbf{Comparison with baselines :} We first compare our method, Multi-modal Patch Association Network ($MMPAN$) where textblocks extracted from the first step are used as input for the second pipeline step, with the baselines. We consider two image semantic segmentation baselines - DeepLabV3+ (DLV3+) \cite{deeplabv3plus2018} and MFCN \cite{colorado}. We train them with an aspect ratio preserving resize of form image to 792x792. For fair comparison with our approach, we use textruns and widgets binary masks as prior by highlighting corresponding regions in a single channel image and concatenate them with the resized form image to obtain the input. For MFCN, loss for different classes are scaled according to pixel area covered by elements of each class (calculated over the dataset) as described in their work. For each baseline, we train a single model\footnote{We also experimented training separate model for each construct for each baseline but it did not provide any improvements} to extract all higher order constructs by processing the penultimate layer output through 3 separate prediction layers - one binary classification layer each for textblocks and choice groups and a ternary classification layer to classify each pixel into field, choice field or background. We report recall and precision for each form structure for different methods in table \ref{Results:pipeline}. For baselines, we match expected groups with segmented outputs through computing Intersection over Union - IoU, keeping a threshold of 0.40 for a correct match. Our method ($MMPAN$) outperforms both the baselines by a very large margin for all the constructs in both precision and recall, even with a much stricter evaluation criteria. 

Figure \ref{visualization} shows structures extracted by our method (top row) and segmentation masks predicted by MFCN (bottom row). It can be seen that MFCN often merges closely spaced textblocks and text fields, is unable to predict complete choice fields and choice groups which might be because it is unable to capture horizontal context. On the other hand, our method is able to disambiguate closely spaced textblocks, extracts all the text fields and choice fields correctly and is able to identify both the choice groups with one wrong hallucination.


\noindent \textbf{Ablation Studies :} To determine the effectiveness of our multi-modal approach, we consider two different variants of $MMPAN$ : 1) $woTE$ where we omit text; 2) $woCE$ where we replace context encoder (CE) Bi-LSTM with an FC that processes spatial and textual representations without taking patch context; 3) $CPAN$ (Convolution Patch Association Network), where the outputs of auxiliary branch of Image Encoder are used to determine final constructs instead of $SAM$; 4) $SPAN$ (Sequence Patch Association Network), where we train the model without visual modality. To do this, we modify the $MMPAN$ by removing IE and FM modules and omit fused representation from the input of $SAM_{LSTM}$. Table \ref{Results:pipeline} summarises the metrics for different constructs for all the variants. It can be seen that $MMPAN$ performs better than both $woTE$ and $woCE$, especially for choice groups where context is most important. For fields, the ablation methods have better precision at the cost of recall while $MMPAN$ has better recall and F-score. $CPAN$ and $SPAN$ perform inferior to $MMPAN$ in both recall and precision for all constructs. In particular, $SPAN$ performs the worst while $CPAN$ performs comparable to $MMPAN$. Comparing $MMPAN$ with $CPAN$ : For textblocks, it achieves $2.33\%$ higher recall with similar precision; for text fields it achieves $3.92\%$ and $1.45\%$ higher recall and precision respectively; for choice fields it shows an improvement of $1.58\%$ and $2.83\%$ in recall and precision respectively while for choice groups, $MMPAN$ performs significantly better with $9.99\%$ higher recall and $13.35\%$ higher precision. We believe this is because choice group is a much larger construct such that association between reference element and candidates in its patch that are part of the same choice group are interdependent. Hence the sequential association decoder is able to capture such dependencies. On the other hand, much inferior performance of $SPAN$ indicates the importance of using visual modality and efficacy of multi-modal fusion approach.

Table \ref{Results:gt} shows results of these variants when tagged textblocks (ground truth) are given as input to the second step of the pipeline instead of textblocks extracted in the first step. Under this setting, we refer to different variants with subscript $GT$. We observe that for each variant, there is a drop in performance (except for textblocks since the first stage remains unaffected). On comparing corresponding rows in table \ref{Results:gt} and table \ref{Results:pipeline}, the drop in performance is most significant for text fields followed by choice group and choice fields. We attribute this behaviour to the fact that incorrectly grouped textblocks in the first step propagates error to the second step. This paves way for further scope of improvement where both the steps are trained together.

\section{Conclusion}
In this paper, we present a novel multi-modal approach for forms structure extraction. Our proposed model uses only lower level elements - textruns and widgets as input to extract larger constructs in a two step pipeline where we extract textblocks first and build upon them with widgets to extract text fields, choice fields and choice groups. Given an element, we train our networks to identify candidate elements in a local neighborhood around it with which it can be grouped using visual, textual and spatial modalities. We establish the efficacy of our multi-modal method by comparing it against individual modality variants. We introduce a stricter and more accurate evaluation criterion (than IoU measures) for evaluating our method. Using this we show that our model gives better results compared to current semantic segmentation approaches evaluated on IoU metric.




 

{\small
\bibliographystyle{ieee}
\bibliography{paper}
}

\end{document}